\pdfoutput=1

\documentclass[11pt]{article}
\usepackage[table]{xcolor}
\usepackage[preprint]{acl}

\usepackage{times}
\usepackage{latexsym}

\usepackage[T1]{fontenc}

\usepackage[utf8]{inputenc}

\usepackage{microtype}

\usepackage{inconsolata}

\usepackage{graphicx}
\usepackage{makecell}

\usepackage{hhline}
\usepackage{longtable}
\usepackage{tablefootnote}
\usepackage{float}

\title{Pipeline Analysis for Developing Instruct LLMs in Low-Resource Languages: A Case Study on Basque}

\author{Ander Corral, Ixak Sarasua \and Xabier Saralegi \\
  Orai NLP Technologies  \\
  \texttt{\{a.corral,i.sarasua,x.saralegi\}@orai.eus}  \\}

\begin{document}
\maketitle
\begin{abstract}
Large language models (LLMs) are typically optimized for resource-rich languages like English, exacerbating the gap between high-resource and underrepresented languages. This work presents a detailed analysis of strategies for developing a model capable of following instructions in a low-resource language, specifically Basque, by focusing on three key stages: pre-training, instruction tuning, and alignment with human preferences. Our findings demonstrate that continual pre-training with a high-quality Basque corpus of around 600 million words improves natural language understanding (NLU) of the foundational model by over 12 points. Moreover, instruction tuning and human preference alignment using automatically translated datasets proved highly effective, resulting in a 24-point improvement in instruction-following performance. The resulting models, Llama-eus-8B and Llama-eus-8B-instruct, establish a new state-of-the-art for Basque in the sub-10B parameter category.
\end{abstract}

\section{Introduction}


Large language models (LLMs) have revolutionized the field of natural language processing (NLP), significantly advancing the state of the art in a wide range of tasks, from language generation to language understanding. Models such GPT-4 \cite{achiam2023gpt} have had a particularly profound impact, showcasing the capabilities of LLMs in real-world applications, showcasing their broad utility across various domains. However, the proprietary nature of these models makes them impractical for many researchers and developers.

\begin{figure}[]
  \includegraphics[width=\linewidth]{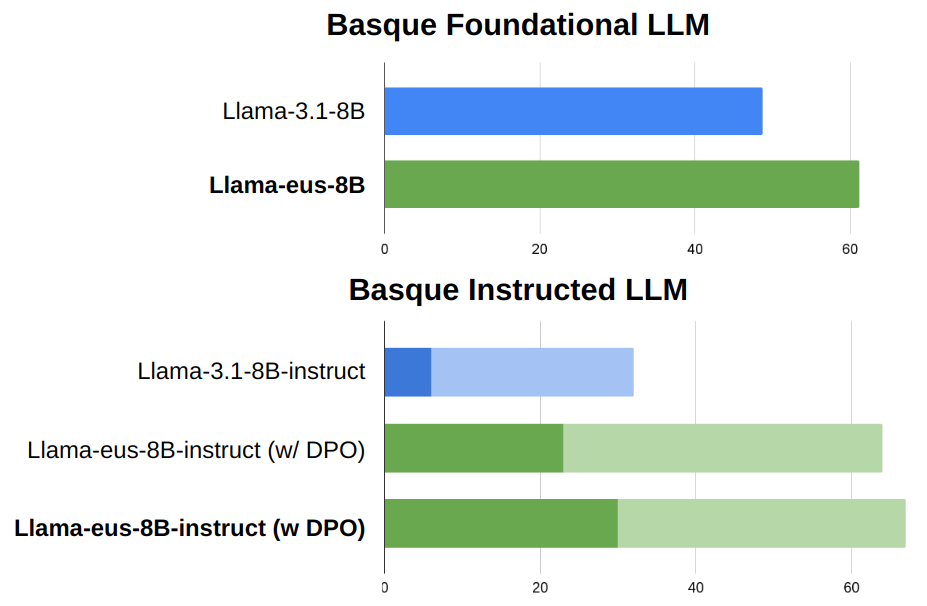}
  \caption{\label{fig:best_results}Comparison of Basque performance between our Llama-eus models and Llama-3.1 baselines. This includes the foundational models' performance on NLU tasks (see Section \ref{sec:cpt}) and the instruction-following performance of instructed models (see Sections \ref{sec:instructing} and \ref{sec:alignment}). In the instructed models, lighter colors indicate partially correct answers.}
\end{figure}

In response, the open-source community has risen to the challenge, developing models that closely rival their proprietary counterparts, such as Llama-3 \cite{dubey2024llama}, Mixtral \cite{jiang2024mixtral}, Qwen2 \cite{yang2024qwen2} or Gemma-2 \cite{team2024gemma}. However, despite these advances, most of these models remain primarily optimized for resource-rich languages—particularly English—which have been trained on vast amounts of data and computational resources. This further widens the gap between high-resource and underrepresented languages, creating a significant barrier to the adoption and effectiveness of LLMs in low-resource languages.

Open-source models, however, offer a unique opportunity to bridge this gap. By leveraging techniques such as transfer learning, it is now feasible to adapt LLMs originally trained on large, predominantly English datasets, and fine-tune them for underrepresented languages using smaller, specialized datasets \cite{cui2023efficient, fujii2024continual, kuulmets-etal-2024-teaching, etxaniz-etal-2024-latxa}. This approach opens the door for developing high-quality LLMs for languages with limited computational resources, ensuring more equitable access to these transformative technologies.

The development of instructed LLMs typically involves three main stages: (a) pre-training a foundational LLM, (b) instructing the LLM, and (c) aligning the model to user preferences. In this paper, we propose strategies to address each of these stages tailored to low-resource languages, and we evaluate the importance of each step in the development of an LLM for such languages. For the purpose of this work, we focus on Basque, a low-resource language as one with minimal representation in the foundational LLM, lacking a sufficiently large and high-quality corpus, and with insufficient data to fully execute the instruction and alignment phases. We conduct experiments using models with fewer than 10 billion parameters, specifically Llama-3.1-8B, to focus on computationally lightweight models.

The main contributions of this work are as follows:
\begin{itemize}
    \item A comprehensive analysis of strategies to implement each stage (pre-training, instruction, and alignment) of the LLM development pipeline, adapted to Basque, a low-resource language.
    
    \item Development of \textbf{Llama-eus-8B}\footnote{Publicly available upon acceptance}, a foundational LLM that achieves the best performance on natural language understanding (NLU) tasks for Basque among sub-10B parameter models. Improvements illustrated in Figure \ref{fig:best_results}.
    
    \item Development of \textbf{Llama-eus-8B-instruct}\footnote{Publicly available upon acceptance}, the first instructed LLM for Basque, achieving state-of-the-art performance among sub-10B parameter models. Improvements illustrated in Figure \ref{fig:best_results}.
    
    \item Introduction of \textbf{new datasets}\footnote{Publicly available upon acceptance} for Basque to support both the evaluation and training of the various phases in the LLM development pipeline.
\end{itemize}

\section{Previous work}

Adapting generative LLMs to other languages has gained significant attention in recent years, with various strategies explored to improve model performance while maintaining computational efficiency. Broadly, the approaches address two main stages: continual pre-training with target language data and instruction tuning to align LLM to user preferences \cite{zhao2024llama}.

Continual pre-training involves further training an LLM on a large corpus of target language textual data. This approach has consistently demonstrated success in boosting language comprehension and generation capabilities \cite{cui2023efficient, fujii2024continual, kuulmets-etal-2024-teaching}. For instance, \newcite{cui2023efficient} report significant improvement in their Chinese-LLaMA over the original LLaMA's \cite{touvron2023llama} ability to understand and generate Chinese content by further pretraining it on extensive Chinese data. MaLA-500 \cite{lin2024mala} is another notable initiative for adapting LLMs to low-resource languages. It utilizes continual pre-training on LLaMA 2 with the Glot500-c dataset \cite{imani2023glot500} to support 534 languages. 


One major challenge in adapting LLMs to new languages is catastrophic forgetting, where a model loses prior knowledge during new language training \cite{zhao2024llama}. To address this, recent studies have incorporated a portion of the original English corpus alongside the target language data during training, as seen in efforts to adapt models to Japanese \cite{fujii2024continual} and Estonian \cite{kuulmets-etal-2024-teaching}.

Instruction fine-tuning has become a vital technique for adapting LLMs to new languages, improving alignment with user preferences and enhancing the model's ability to follow instructions. For instance, studies by \newcite{cui2023efficient} and \newcite{jiang2024mixtral} illustrate how instruction tuning refines task-specific instruction adherence in Chinese. Their findings indicate that when combined with continual pre-training, instruction tuning significantly boosts performance, especially for complex language tasks. Additionally, \newcite{jiang2024mixtral} demonstrated that initiating the process with the foundational model, rather than the instruction model, is more effective for transferring language abilities.


In the context of Basque, the Latxa \cite{etxaniz-etal-2024-latxa} family of foundational models,which ranges from 7 to 70 billion parameters and is based on LLaMA 2, represents a significant advancement in adapting LLMs to the Basque language. Through continual pre-training on a specialized Basque corpus, these models have achieved substantial improvements in processing Basque text.


\section{Adapting a Foundational Model to Basque}
\label{sec:cpt}

In this section, we focus on adapting a foundational English-centric model to Basque by continual pre-training with high-quality Basque data. Continual pre-training is a crucial technique for adapting large language models to new languages, where the goal is to incrementally refine the model's capabilities for the new language. 

\subsection{Foundational Model Choice: Balancing Performance and Efficiency}

We chose Llama-3.1-8B \cite{dubey2024llama} as our base foundational model for this work. Although larger models, such as the 70B version, were initially considered and demonstrated superior capabilities, their high computational and memory demands pose significant challenges in resource-limited environments, both during training and deployment, making Llama-3.1-8B the optimal choice for our scenario. 

We initially evaluated earlier versions, such as Llama-2 \cite{touvron2023llama}, but ultimately selected Llama-3.1 as it consistently achieved the best results for Basque NLU tasks among all Llama variants. In addition to its superior results, Llama-3.1 features an expanded vocabulary designed to better support multiple languages and offers broader native support for a wider set of languages. These qualities made Llama-3.1 the optimal choice for our model adaptation to Basque. See Appendix \ref{app:base_model_evaluation} for the evaluation results.

\subsection{Training Data}
\label{sec:training-data}

For continual pre-training, we utilized the \textbf{ZelaiHandi} dataset \cite{ZelaiHandi}, the largest collection of freely licensed and clean Basque texts available as of October 2024. This dataset, comprising approximately 521 million words (around 1.5B tokens with Llama-3.1 tokenizer), was meticulously compiled from selected web sources, ensuring that only high quality documents published under free license were included. By high quality we refer to texts that are well-formed, free of excessive noise or errors, and representative of formal and diverse language use across various domains (see Appendix \ref{app:cpt_training_data} for further details on the sources considered during ZelaiHandi creation). 

Compared to other existing datasets for Basque (see Appendix \ref{app:cpt_results_different_datasets}), ZelaiHandi proved to be the most favourable choice, combining high-quality content with an open license. Notably, it demonstrates competitive performance with significantly less data, contributing to more computationally efficient pre-training compared to larger datasets.

To further enhance the continual pre-training process, we incorporated the \textbf{FineWeb} dataset \cite{penedo2024finewebdatasetsdecantingweb}, which comprises over 15 trillion tokens of cleaned and deduplicated English web data sourced from CommonCrawl and licensed under ODC-By 1.0 license. For our purposes, we used a random subset of approximately 300 million tokens. As observed in recent literature \cite{kuulmets-etal-2024-teaching, etxaniz-etal-2024-latxa, fujii2024continual}, this approach aims to avoid catastrophic forgetting of previously learned competencies in English, which hinders transfer-learning from English. The objective is to develop formal linguistic competencies for Basque, including grammar and vocabulary, while leveraging the functional linguistic skills—such as reasoning and world knowledge—acquired from the English data during the original training of Llama 3.1.

\subsection{Evaluation benchmarks}

To evaluate our model's performance in Basque, we employed a variety of benchmarks, including both manually translated well established English benchmarks and existing Basque benchmarks. This comprehensive evaluation approach allows us to measure the model’s capabilities across different tasks, ensuring a robust understanding of its formal and functional competencies in the Basque language.

We manually created four new benchmarks for Basque by translating samples from well-established English benchmarks:

\begin{itemize}
    \item \textbf{ARC\_HT\_eu\_sample}: A subset of 250 samples manually translated to Basque from the ARC dataset \cite{allenai:arc}. The ARC dataset consists of genuine grade-school level, multiple-choice science questions.
    \item \textbf{Winogrande\_HT\_eu\_sample}: A subset of 250 samples manually translated to Basque from the WinoGrande dataset \cite{ai2:winogrande}. WinoGrande is a dataset of 44k problems specifically designed to test commonsense reasoning.
    \item \textbf{MMLU\_HT\_eu\_sample}: A subset of 270 samples manually translated to Basque from the MMLU dataset \cite{hendryckstest2021}. The MMLU dataset is a massive multitask test consisting of multiple-choice questions from various branches of knowledge. The test spans subjects in the humanities, social sciences, hard sciences, and other areas that are important for some people to learn.    
    \item \textbf{HellaSwag\_HT\_eu\_sample}: A subset of 250 samples manually translated to Basque from the HellaSwag dataset \cite{zellers2019hellaswag}. The HellaSwag dataset commonsense NLI evaluation benchmark.
\end{itemize}

All benchmarks were translated by a native Basque speaker. For all newly created benchmarks, we also provide the corresponding English samples, enabling a direct comparison of model performance between Basque and English. This allows for a clear assessment of the performance gap between the model's competencies in English and Basque.

In addition to those new benchmarks, we leveraged existing Basque benchmarks:

\begin{itemize}
    \item \textbf{BL2MP} \cite{urbizu-etal-2024-well}: The BL2MP test set, designed to assess the grammatical knowledge of language models in the Basque language, inspired by the BLiMP \cite{warstadt-etal-2020-blimp-benchmark} benchmark.
    \item \textbf{BasqueGLUE} \cite{urbizu2022basqueglue}: BasqueGLUE is a NLU benchmark for Basque, which has been elaborated from previously existing datasets and following similar criteria to those used for the construction of GLUE and SuperGLUE.
    \item \textbf{Belebele} \cite{bandarkar-etal-2024-belebele}: Belebele is a multiple-choice machine reading comprehension dataset spanning 122 language variants.
    \item \textbf{X-StoryCloze} \cite{xstorycloze}: XStoryCloze consists of the professionally translated version of the English StoryCloze dataset to 10 non-English languages. It is a commonsense reasoning framework for evaluating story understanding, story generation, and script learning
    \item \textbf{EusProficiency}, \textbf{EusReading}, \textbf{EusExams}, and \textbf{EusTrivia} \cite{etxaniz-etal-2024-latxa}: Basque-specific benchmarks covering proficiency tests based on past EGA exams (C1 level Basque), reading comprehension, public service exam preparation, and trivia questions respectively.
\end{itemize}

Model evaluations were conducted with the LM Evaluation Harness library \cite{eval-harness} by Eleuther AI. Models accuracies are evaluated on tasks following an in-context few-shot fashion where most of the tasks are evaluated by using 5 in-context examples, except for HellaSwag\_HT\_eu\_sample (10-shot), ARC\_HT\_eu\_sample (25-shot), BL2MP (0-shot), X-StoryCloze (0-shot) and EusReading (1-shot). The few-shot choice is made based on the Open LLM Leaderboard \cite{open-llm-leaderboard}.

\subsection{Training setup}

We opted for mixing Basque and English data with a ratio of 80:20 in order to ensure that the model's competence in Basque improves without suffering from catastrophic forgetting that hinders transfer-learning from English \cite{kuulmets-etal-2024-teaching}. To prevent language interference during the mixing of languages, we implemented a modified sequence packing strategy, ensuring that only examples from the same language were packed together in a single sequence. 

We conducted full fine-tuning of all model parameters to maximize the model’s ability to learn linguistic nuances in Basque. We utilized the Hugging Face Transformers \cite{wolf-etal-2020-transformers} library, alongside DeepSpeed ZeRO \cite{rajbhandari2020zero} and Accelerate \cite{accelerate}, to manage efficient large-scale training.

Training was carried out on 8 NVIDIA A100 80GB GPUs over 4 epochs, with a sequence length of 4096 tokens and an effective batch size of approximately 2 million tokens. In total, 7.2 billion tokens were processed, with training guided by a cosine learning rate schedule, peaking at 1e-4, and a warm-up phase comprising 10\% of the total steps. All remaining hyperparameters followed the configurations established by \newcite{dubey2024llama}. Estimated carbon emissions are detailed in Appendix \ref{app:carbon-emissions}.

\subsection{Results}

To assess the validity of our approach, we compare our model against various versions of \textbf{Llama-3.1}, specifically the 8B and 70B models. The Llama-3.1-8B model serves as the base for our continual pre-training, establishing a baseline for evaluating the enhancements achieved through our method. Additionally, we include the \textbf{Latxa} models \cite{etxaniz-etal-2024-latxa} in our comparison, as they represent another open-source family of large language models specifically adapted to Basque, with parameter sizes ranging from 7 billion to 70 billion. As the only existing models tailored for the Basque language, Latxa provides a crucial baseline for evaluating our results.

We categorize the evaluation into sub-10 billion and over-10 billion parameter models to gain a clearer understanding of the performance differences across various model sizes. This distinction enables a fairer comparison of our model against both smaller and larger-scale architectures. 


\begin{table*}[]
\centering
\begin{tabular}{l|ccc|ccc}
\hline
\textbf{Benchmark} & \makecell{\textbf{Latxa v1.2} \\ \textbf{7B}} & \makecell{\textbf{Llama 3.1} \\ \textbf{8B}} & \makecell{\textbf{Llama-eus} \\ \textbf{8B}} & \makecell{\textbf{Latxa v1.2} \\ \textbf{13B}} & \makecell{\textbf{Latxa v1.2} \\ \textbf{70B}} & \makecell{\textbf{Llama 3.1} \\ \textbf{70B}} \\ \hline
ARC\_eu              & 54.80  & 42.80  & \textbf{55.20}    & 55.60  & 64.80  & \underline{\textbf{67.20}}                   \\ 
Winogrande\_eu       & 65.60  & 56.80  & \textbf{67.20}    & 69.60  & \underline{\textbf{72.80}} & 70.00                   \\ 
MMLU\_eu             & 34.44   & 48.52 & \cellcolor[HTML]{9BC38C}\textbf{53.33} & 39.63  & 47.78  & \underline{\textbf{63.70}}                   \\ 
HellaSwag\_eu        & 61.20 & 46.80 & \cellcolor[HTML]{ADDB9B}\textbf{63.60}  & 61.60  & \underline{\textbf{67.20}} & 63.60                   \\ \hline
BL2MP                & \underline{\textbf{89.33}}  & 60.50  & \cellcolor[HTML]{9BC38C}89.22  & 88.67 & \textbf{88.72}   & 67.89                   \\ 
Belebele             & 37.33 & 61.78  & \cellcolor[HTML]{9BC38C}\textbf{73.44 } & 53.89 & 71.67 & \underline{\textbf{87.67}}                   \\ 
X-StoryCloze         & 65.45 & 55.66 & \textbf{65.72}  & 66.51 & \underline{\textbf{70.55}}  & 65.98                   \\ 
EusExams             & 33.82 & 45.65 & \cellcolor[HTML]{9BC38C}\textbf{52.51} & 43.66  & 51.90  & \underline{\textbf{64.62}}                  \\ 
EusProficiency       & 30.26  & 32.50 & \cellcolor[HTML]{ADDB9B}\textbf{48.44} & 44.11 & \underline{\textbf{60.65}} & 44.86                   \\ 
EusReading           & 26.99  & 43.18 & \cellcolor[HTML]{9BC38C}\textbf{54.55}  & 34.94  & 52.27  & \underline{\textbf{72.44}}                  \\ 
EusTrivia            & 42.16 & 44.49 & \textbf{56.21} & 56.38 & \underline{\textbf{62.45}} & 60.23                   \\ 
BasqueGLUE           & 52.56  & 46.33  & \cellcolor[HTML]{ADDB9B}\textbf{55.27} & 53.36 & 59.74 & \underline{\textbf{63.50}}                   \\ \hline
\rowcolor[HTML]{E6E6E6}\textbf{Average}  & 49.50 & 48.75 & \cellcolor[HTML]{ADDB9B}\textbf{61.22}  & 55.66 & 64.21 & \underline{\textbf{65.97}}          \\ \hline
\end{tabular}
\caption{\label{tab:cpt_results_basque}Basque evaluation results of models with fewer than 10B parameters and more than 10B parameters. \textbf{Bold} highlights the best results among models classified according to parameter counts, while the \underline{underlined} value denotes the overall best result across all configurations. The light green indicates that Llama-eus-8B surpasses the Latxa-13B model, while the dark green indicates that it also the Latxa-70B model.}
\end{table*}

In the sub-10B parameter category, the results demonstrate that Llama-eus-8B significantly outperforms all other models across the Basque benchmarks (see Table \ref{tab:cpt_results_basque}), with the exception of a minor decline in the BL2MP task. Llama-eus-8B achieves the best average score, 61.22, which is notably higher than the 49.50 recorded by Latxa v1.2 7B and the 48.75 achieved by Llama-3.1-8B. This represents an average improvement of 12.47 points over the base model Llama-3.1-8B, underscoring the effectiveness of our continual pre-training strategy that incorporates Basque data. 

When comparing our Llama-eus-8B model with those in the over-10B parameter category, it is noteworthy that Llama-eus-8B not only surpasses Latxa-13B but also competes effectively against Latxa-70B across 5 out of 12 Basque benchmarks (see Table \ref{tab:cpt_results_basque}). While Latxa-70B excels in certain categories, Llama-eus-8B achieves an impressive average score of 61.22, trailing only by 3 points behind Latxa-70B, despite having significantly fewer parameters. This indicates a favorable balance between model size and performance, showcasing Llama-eus-8B's ability to deliver solid results without the need for a larger model. Interestingly, Llama-3.1-70B achieves the highest average score, 65.97, across the Basque benchmarks, even though it has not been specifically trained for Basque tasks.

We also assessed the English performance of our Llama-eus-8B model following the continual pre-training phase, as maintaining its initial competencies is crucial. The analysis revealed that the model experiences only a modest decrease of 1.96 points in average English scores compared to the baseline Llama-3.1-8B. This outcome indicates that while Llama-eus-8B has been effectively adapted for Basque, it continues to demonstrate a commendable level of competency in English, preserving its foundational knowledge. However, a significant performance gap, 13.28 points, remains between Basque and English across all evaluated models. For additional details and insights into the results, readers are encouraged to consult Appendix \ref{app:english_performance}.


\section{Instruction Tuning the Model in Basque}
\label{sec:instructing}

In this section, we present experiments aimed at enabling a foundational LLM to follow instructions in Basque. Our focus is twofold: first, to assess whether using a foundational model specifically adapted to Basque, as described in the previous chapter, offers any advantages; and second, to evaluate the feasibility of leveraging English instruction datasets translated into Basque via automatic translation for the instruction fine-tuning process.

\subsection{Instruction datasets for Basque}
\label{sec:ift_datasets}

Given the limited availability of instruction datasets in Basque and the cost challenges of manually creating equivalents to those in English, we explore the generation of Basque instruction datasets through automatic translation.

For this analysis, we utilize two widely recognized English instruction datasets, No\_Robots \cite{no_robots} and SlimOrca \cite{OpenOrca}. No\_Robots is a smaller, manually curated dataset licensed under CC BY-NC 4.0 license. It contains 9,500 instructions covering various tasks, including generation, open QA, and brainstorming. In contrast, SlimOrca is significantly larger, featuring 517,982 automatically generated instructions with GPT-4 \cite{achiam2023gpt}. It is distributed under MIT License.

To produce the Basque counterparts of these datasets, referred to as \textbf{No\_Robots\_eu} and \textbf{SlimOrca\_eu}, we employ the Elia machine translation platform\footnote{https://elia.eus/translator}, which achieves a translation quality of 19.3 BLEU and 52.2 chrF++ for the English-to-Basque direction on the FLORES-200 Evaluation Benchmark \cite{nllb2022}. Details of the newly generated datasets are presented in Table \ref{tab:instruction_data}.

\begin{table}[]
\centering
\begin{tabular}{lcc}
\hline
& \textbf{\# Instructions} & \textbf{Avg. words} \\ \hline
No\_Robots\_eu  & 9.5K          &  157.9   \\
SlimOrca\_eu   & 518K          &  227.8   \\
\hline
\end{tabular}
\caption{\label{tab:instruction_data}Summary of Basque instruction datasets created through machine translation from English No\_Robots and SlimOrca datasets, detailing the number of instructions and the average word count for each dataset.}
\end{table}

\subsection{Evaluation methodology}

Although manual evaluation requires significant effort, we opted for this approach to assess the ability of LLMs to follow instructions, given the limitations of automatic evaluation methods in language generation tasks. Specifically, a native Basque speaker evaluated the models using a random sample of 100 instructions from the No\_Robots test set, which comprises 500 instructions across 10 categories. The random selection ensured a minimum representation from each category, with the ‘coding’ category excluded to focus solely on text-based tasks. The 100 instructions were manually translated into Basque. Additional details on the sampling process are provided in Appendix \ref{app:instruction_datasets}.

To generate model responses, we employed greedy search decoding during inference to guarantee both stability and reproducibility. The generated outputs were manually classified into three categories: a) correct, b) partially correct, and c) wrong. A response was deemed correct if it fully addressed the task without introducing hallucinations or misinformation; partially correct if it accomplished part of the task but contained inaccuracies or incomplete elements; and wrong if it failed to satisfy the task requirements.

\subsection{Training setup}

The instruction fine-tuning of the foundational models was performed using LoRA \cite{LoRA}, as preliminary experiments demonstrated that this approach yielded better results than full fine-tuning. The fine-tuning process employed a batch size of 64 instructions and a cosine learning rate scheduler with a peak learning rate of 2e-5. The LoRA-specific hyperparameters were set to a rank of 64, an alpha value of 16, and a dropout probability of 0.1. All other hyperparameters remained consistent with those used during the pretraining phase.

\subsection{Results}

\begin{table}[]
    \centering
    \begin{tabular}{lccc}
        \hline
        \textbf{Model} &  \textbf{Corr.}  &  \makecell{\textbf{Partial} \\ \textbf{corr.}}  &  \textbf{Wrong}  \\ \hline
        Llama-eus-8B  \\ 
        \textit{ +No\_Robots\_eu} &  15\%  &  33\%  &  52\%  \\
        \textit{ +SlimOrca\_eu} &  \textbf{23\%}  &  \textbf{41\%}  &  \textbf{36\%}  \\ \hline
        Llama-3.1-8B \\
        \textit{ +SlimOrca\_eu} &  14\%  &  34\%  &  52\%  \\ \hline
        
        Llama-3.1-8B-it. &  6\%  &  26\%  &  68\%  \\
        \hline
    \end{tabular}
    \caption{Manual evaluation results of the instruction tuned models for Basque using different automatically translated intructions datasets. \textit{Llama-3.1-8B-it.} refers to the original Llama-3.1-8B-instruct model. In \textbf{bold} we highlight the best performing model.}
    \label{tab:ift_results}
\end{table}

\textbf{Can Translated Data Effectively Instruct Basque LLMs?} We evaluated the feasibility of instructing the Basque-adapted foundational model, Llama-eus-8B, using the machine-translated datasets No\_Robots\_eu and SlimOrca\_eu (presented in Section \ref{sec:ift_datasets}). The results presented in Table \ref{tab:ift_results} demonstrate a significant performance improvement over the Llama-3.1-8B-instruct baseline, previously the best 8B-instructed model for Basque. The Llama-eus-8B model trained with No\_Robots\_eu achieves a 9 percentage point increase in the correct rate and a 7 percentage point increase in the partially correct rate compared to Llama-3.1-8B-instruct. The gains are even more pronounced for Llama-eus-8B trained on SlimOrca\_eu, which shows enhancements of 17 points in the correct rate and 15 points in the partially correct rate relative to Llama-3.1-8B-instruct. In addition, the results indicate that SlimOrca\_eu is more suitable for intruction tunining, as the model tuned with SlimOrca\_eu surpasses its counterpart by 8 points. This indicates that higher quality of No\_Robots\_eu does not offset its smaller size compared to SlimOrca\_eu.

\textbf{How Does Pre-training on Basque Impact Instruction Tuning Performance?} We assessed the advantage of using a foundational model adapted to Basque language versus one without specific adaptation (Llama-eus-8B vs. Llama-3.1-8B). We compare the performance of the Llama-eus-8B + SlimOrca\_eu model to that of the Llama-3.1-8B + SlimOrca\_eu. The results, shown in Table \ref{tab:ift_results}, demonstrate the superior performance of Llama-eus-8B + SlimOrca\_eu, with an improvement of 9 points in the correct rate and 7 points in the partially correct rate. This highlights the significant benefits of instructing with a foundational model that has been specifically adapted to the target language.

The top-performing model trained with SlimOrca\_eu will henceforth be referred to as \textbf{Llama-eus-8B-instruct}.

\section{Alignment to Human Preferences in Basque}
\label{sec:alignment}

In this section, we present experiments focused on adapting instruction tuned models for Basque to align with human preferences, with the goal of improving their ability to generate answers in Basque. Similar to the experiments described in Section \ref{sec:instructing}, we address two key aspects: first, the feasibility of achieving alignment using preference sets in Basque, generated through the automatic translation of English datasets; and second, the potential benefits of adapting an instructed model specifically to Basque, compared to using a general instructed model without explicit adaptation to the language.

Several algorithms have been developed to improve the alignment of language model responses with human preferences. Among these, Direct Preference Optimization (DPO) \cite{DPO, dubey2024llama} has emerged as a particularly promising approach due to its simplicity and effectiveness. DPO algorithm requires a dataset consisting of paired samples of "preferred" and "rejected" reponses to a given prompt. DPO leverages these preference pairs to directly optimize the model’s output generation, focusing on improving the likelihood of the preferred responses compared to the rejected ones. Unlike traditional reinforcement learning methods like PPO \cite{PPO}, DPO bypasses the need for reward models, simplifying the training process by operating directly on these preference rankings. Consequently, we have chosen to implement this algorithm in our study. 

\subsection{Preference Dataset for Basque}

Achieving significant improvements with human preference adaptation algorithms typically requires a large training dataset containing thousands of examples, ideally constructed from manual human feedback on LLM outputs. However, this process is costly, so alternatives have been explored, such as using LLMs as evaluators or rankers to reduce the need for extensive manual input \cite{open_hermes_preferences, starling2023}.

Since no such dataset exists in Basque, we opted to translate an existing public dataset from English to Basque. After evaluating several options—OpenHermes \cite{open_hermes_preferences}, UltraFeedback \cite{cui2023ultrafeedback}, and Nectar \cite{starling2023}—we selected UltraFeedback as the most suitable for our experiments. UltraFeedback is a large-scale preference dataset consisting of almost 64k samples generated by various LLMs, with GPT-4 annotations covering four key aspects—instruction-following, truthfulness, honesty, and helpfulness—based on prompts from diverse sources. It is distributed under MIT License. While it is the smallest of the options, it offers an ideal balance between quality and size, providing enough data for effective experimentation without overwhelming computational resources.

We followed the same methodology as outlined for the instruction datasets (see Section \ref{sec:ift_datasets}) to translate the dataset to Basque. The resulting dataset, named \textbf{UltraFeedback\_eu}, contains 61,135 triplets, each comprising a prompt, a preferred response, and a rejected response. An example is provided in Appendix \ref{app:preferences_datasets}.

\subsection{Experimentation Results}

\begin{table}[]
    \centering
    \begin{tabular}{lccc}
        \hline
        \textbf{Model} &  \textbf{Corr.}  &  \makecell{\textbf{Partial} \\ \textbf{corr.}}  &  \textbf{Wrong}  \\ \hline
        Llama-eus-8B-it. &  23\%  &  \textbf{41\%}  &  36\%  \\
        \textit{ +UFeedback\_eu} &  \textbf{30\%} &  37\%  &  \textbf{33\%}  \\ \hline
        Llama-3.1-8B-it. &  6\%  &  26\%  &  68\%  \\
        \textit{ +UFeedback\_eu} &  2\%  &  25\%  &  73\% \\ \hline
    \end{tabular}
    \caption{Manual evaluation results of the models aligned to human preferences for Basque using UltraFeedback\_eu, an automatically translated feedback dataset. \textit{Llama-3.1-8B-it.} refers to the original Llama-3.1-8B-instruct model. In \textbf{bold} we highlight the best performing model.}
    \label{tab:dpo_results}
\end{table}

\textbf{Is translated preference data feasible for alignment?} To evaluate the effectiveness of the UltraFeedback\_eu dataset, we conducted DPO training on the best performing model for Basque from Section \ref{sec:instructing}, Llama-eus-8B-instruct, an instruction-tuned model trained on SlimOrca\_eu. Manual evaluation was carried out using the same test set employed in the instruction-following experiments described in Section \ref{sec:instructing}. The results, shown in Table \ref{tab:dpo_results}, indicate that model following DPO achieved a notable 7-point improvement in the accuracy of correct answers over Llama-eus-8B-instruct. These findings demonstrate the viability of using machine-translated preference data for alignment. This model will henceforth be referred to as \textbf{Llama-eus-8B-instruct-DPO}.

\textbf{Choosing the right base model for alignment: Basque vs. English} To evaluate the benefit of using as a base model a model explicitly adapted to Basque, we compare the performance of our Llama-eus-8B-instruct-DPO model against Llama-3.1-8B-instruct aligned with UltraFeedback\_eu. As shown in Table \ref{tab:dpo_results}, the results of the manual evaluation highlight the clear superiority of Llama-eus-8B-instruct-DPO. Notably, the model based on Llama-3.1-8B-instruct performs worse than its base version. This demonstrates the substantial advantage of starting with a model explicitly tailored to the target language.

\textbf{English performance.} To evaluate the performance gap between our best Basque-adapted model, Llama-eus-8B-instruct-DPO, and state-of-the-art instruction-tuned models in English, we assessed the Llama-3.1-8B-instruct model using the English version of the 100-sample test set derived from the No\_Robots dataset. The results revealed 91 correct responses, 6 partially correct, and 3 incorrect, highlighting the potential for further improvement in models specifically trained for Basque.

\section{Conclusions}

This work provides a comprehensive analysis of strategies for developping a model capable of following instructions in a low-resource language such as Basque, focusing on three key stages: pre-training, instruction tuning, and alignment with human preferences. The results demonstrate that tailoring each of these stages to the target language significantly enhances model performance compared to baseline English-centric models based on Llama 3.1.

In the pre-training stage, it has been shown that continual pre-training of a foundational English-centric LLM with a high-quality corpus of fewer than 1 billion words in the low-resource language can yield an average improvement of over 12 points in natural language understanding (NLU) tasks.

In the instruction tuning and human preference alignment stages, automatic translation of English datasets proved effective for training models to follow instructions in Basque, surpassing the instructed Llama 3.1 baseline. Using Basque-adapted models as the training base further enhanced performance. Following instruction tuning, the application of DPO with translated preference datasets improved model accuracy by up to 7 points.  

The experimental results establish Llama-eus-8B as the most suitable foundational model for Basque within its parameter scale, and Llama-eus-8B-instruct as the first model specifically instruction-tuned for Basque, achieving the highest performance in this language. However, despite these advances, the performance of both models in Basque still lags behind Llama 3.1's performance in English, underscoring a substantial performance gap and indicating considerable room for further enhancement. Especially Llama-eus-8B-instruct, still require further refinement to become competitive in real-world production environments.

\section*{Limitations}

The experimentation focused on strategies to develop a model capable of following instructions in a low-resource language like Basque, using Llama-3.1-8B as the baseline within the sub-10B parameter category. We acknowledge that results may vary across different architectures and model sizes, and thus the findings of this study may not be directly applicable to other LLMs with differing characteristics. Similarly, the results for Basque may not necessarily be replicable for other low-resource languages, as capabilities can vary across linguistic contexts.

The evaluation of the models’ ability to follow instructions was carried out manually, using a sample of 100 instructions randomly selected from the No\_Robots dataset, a high-quality test set manually curated. We chose to evaluate several models with a sample of 100 examples rather than fewer models with a larger sample. The results of this manual evaluation and the conclusions drawn should be interpreted within the context of the nature and size of this test set.

In the analysis of the pipeline for developing adapted LLMs for low-resource languages, we investigated strategies that outperformed the proposed baselines. However, these strategies may be suboptimal. This research represents a solid first step in developing techniques for building fine-tuned LLMs for Basque.

\section*{Ethics Statement}

Like other generative language models, Llama-eus-8B and Llama-eus-8B-instruct may produce information that does not align with certain ethical values, such as displaying negative social biases towards some minorities. Although the pre-training of these models was carried out using a corpus based on reliable sources, there is a possibility that unwanted biases or other ethical patterns were learned. Plans are in place to correct the identified social biases in these models in the short term. In any case, we recommend that Llama-eus-8B and Llama-eus-8B-instruct be used in controlled environments, preceded by a thorough analysis of any potential harm they could cause in the specific use cases for which they are employed.

\section*{Acknowledgments}

This work is part of the BasqueLLM project, titled "First steps towards an artificial intelligence in Basque based on LLMs" (EXP: 2023-CIEN-000081-01), partially funded by the Guipuzcoa Science, Technology and Innovation Network Program of the Provincial Council of Gipuzkoa. Model training and development were conducted using the Hyperion system at the Donostia International Physics Center (DIPC).

\bibliography{custom}

\appendix

\section{Foundational Model Choice: Balancing Performance and Efficiency}

We chose Llama-3.1-8B \cite{dubey2024llama} as our base foundational model for this work. Although larger models, such as the 70B version, were initially considered and demonstrated superior capabilities, their high computational and memory demands pose significant challenges in resource-limited environments, both during training and deployment, making Llama-3.1-8B the optimal choice for our scenario. 

We initially evaluated earlier versions, such as Llama-2 \cite{touvron2023llama}, but ultimately selected Llama-3.1 as it consistently achieved the best results for Basque NLU tasks among all Llama variants. In addition to its superior results, Llama-3.1 features an expanded vocabulary designed to better support multiple languages and offers broader native support for a wider set of languages. These qualities made Llama-3.1 the optimal choice for our model adaptation to Basque. In Table \ref{tab:base_model_evaluation} we report different base models comparison results, comprising Llama-2-7B, Llama-3-8B and Llama-3.1-8B.

\label{app:base_model_evaluation}

\begin{table*}[]
\centering
\begin{tabular}{l|ccc}
\hline
\textbf{Benchmark} & \textbf{Llama 2 7B} & \textbf{Llama 3 8B} & \textbf{Llama 3.1 8B} \\ \hline
ARC\_eu & 22.40 & \textbf{43.60} & 42.80 \\
Winogrande\_eu & 44.80 & 54.40 & \textbf{56.80} \\
MMLU\_eu & 28.52 & 44.07 & \textbf{48.52} \\
HellaSwag\_eu & 28.80 & 44.40 & \textbf{46.80} \\ \hline
BL2MP & 55.94 & 58.28 & \textbf{60.50} \\
Belebele & 29.00 & 60.11 & \textbf{61.78} \\
X-StoryCloze & 50.56 & \textbf{55.86} & 55.66 \\
EusExams & 28.87 & 44.45 & \textbf{45.65} \\
EusProficiency & 23.33 & 31.96 & \textbf{32.50} \\
EusReading & 24.72 & 41.76 & \textbf{43.18} \\
EusTrivia & 30.38 & \textbf{45.83} & 44.49 \\
BasqueGLUE & 38.15 & 45.67 & \textbf{46.33} \\ \hline
Average & 32.81 & 47.53 & \textbf{48.75} \\ \hline
\end{tabular}
\caption{\label{tab:base_model_evaluation}Evaluation of different open-source models.}
\end{table*}

\section{ZelaiHandi Dataset Information}
\label{app:cpt_training_data}

For continual pre-training, we utilized the ZelaiHandi dataset \cite{ZelaiHandi}, the largest collection of freely licensed and clean Basque texts available as of October 2024. This dataset, comprising approximately 521 million words (around 1.5B tokens with Llama-3.1 tokenizer), was meticulously compiled from selected web sources, ensuring that only high quality documents published under free license were included. By high quality we refer to texts that are well-formed, free of excessive noise or errors, and representative of formal and diverse language use across various domains. Table \ref{tab:zelaihandi_stats} summarizes corpus statistics and license information. This commitment to quality and accessibility makes ZelaiHandi particularly valuable for effective model training.

\begin{table*}[]
\centering
\begin{tabular}{lllll}
\hline
source & domain & tokens (M) & license \\
\hline
Tokikom\tablefootnote{https://tokikom.eus/bazkideak/} & news & 163.72 & cc-by / cc-by-sa \\
Berria\tablefootnote{https://berria.eus} & news & 125.72 & cc-by-sa 4.0 \\
Eusko Legebiltzarra\tablefootnote{https://www.legebiltzarra.eus} & administrative & 80.16 & public domain \\
Wikipedia\tablefootnote{https://wikipedia.org} & wikipedia & 63.83 & cc-by-sa 4.0 \\
Argia\tablefootnote{https://www.argia.eus} & news & 17.44 & cc-by-sa 4.0 \\
Addi\tablefootnote{https://addi.ehu.es/} & science & 17.09 & several cc variants \\
Opendata Euskadi subtitles\tablefootnote{https://opendata.euskadi.eus/catalogo/-/euskarazko-azpitituluak-filmak-eta-telesailak-euskaraz/} & subtitles & 11.30 & cc-by-sa \\
Hitza\tablefootnote{https://hitza.eus} & news & 9.57 & cc-by-sa 4.0 \\
GFA\tablefootnote{https://www.bngipuzkoa.eus/} & administrative & 8.46 & Custom copyright license \\
Zientzia.eus\tablefootnote{https://zientzia.eus} & science & 8.37 & cc-by-sa \\
Susa\tablefootnote{https://www.susa-literatura.eus/} & literature & 5.77 & cc-by \\
BFA\tablefootnote{https://jjggbizkaia.eus} & administrative & 4.26 & Custom copyright license \\
Zientzia Kaiera\tablefootnote{https://zientziakaiera.eus/} & science & 1.94 & cc-by-sa 3.0 \\
Ekaia\tablefootnote{https://ojs.ehu.eus/index.php/ekaia} & science & 1.80 & cc-by-nc-nd 4.0 \\
Ikergazte\tablefootnote{https://www.buruxkak.eus/bilatu?bilaketa=ikergazte} & science & 1.68 & cc-by-sa 3.0 \\
Game-erauntsia\tablefootnote{https://gamerauntsia.eus/} & videogames & 0.40 & cc-by-sa 4.0 \\ \hline
\textbf{Total} &  &  & \textbf{521.55} & \\ \hline
\end{tabular}
\caption{\label{tab:zelaihandi_stats}ZelaiHandi corpus statistics and license information.}
\end{table*}

\section{Performance Analysis of Dataset Choice for Continual Pre-training}
\label{app:cpt_results_different_datasets}

Before the continual pre-training phase, we conducted an analysis to determine the most adequate Basque dataset for the task. We focused on analyzing the trade-offs between dataset size, quality, and their impact on computational efficiency. To that end, apart from the \textbf{ZelaiHandi} dataset (see Section \ref{sec:training-data}) we considered two additional Basque datasets to evaluate their impact on model performance:

\begin{itemize}
    \item \textbf{ZelaiItxi}: a proprietary dataset that extends ZelaiHandi by incorporating additional closed-source data from selected sources of comparable quality.
    \item \textbf{Latxa} \cite{etxaniz-etal-2024-latxa}: the dataset used to trained the Latxa \cite{etxaniz-etal-2024-latxa} family of large language models for Basque. 
\end{itemize}

\begin{table}[]
\centering
\begin{tabular}{llll}
\hline
           & \textbf{Docs} & \textbf{Words} & \textbf{Tokens} \\ \hline
ZelaiHandi & 1.61M          & 512M           & 1.55B            \\
ZelaiItxi  & 2.86M            & 692M           & 2.51B              \\ \hline
Latxa      & 4.30M         & 1.22B          & 3.46B             \\ \hline
\end{tabular}
\caption{\label{tab:datasets}Sizes in terms of the number of documents, words and tokens (Llama 3.1 tokenizer) for the Basque datasets.}
\end{table}

Table \ref{tab:datasets} presents the dataset sizes in terms of the number of documents, words, and tokens (using the Llama 3.1 tokenizer). 

We conducted continual pre-training experiments with various data configurations to assess their impact on model adaptation and training efficiency. The configurations included:

\begin{itemize} 
\item \textbf{ZelaiHandi}: combining Basque data from ZelaiHandi with English data from FineWeb to investigate the effects of bilingual training on model performance.
\item \textbf{ZelaiItxi}: By including additional high-quality proprietary sources from ZelaiItxi, we aimed to analyze how enhancing the training set with curated data influences the model's adaptation. English data is also included.
\item \textbf{Largest open-source dataset}: This configuration combined ZelaiHandi with Latxa, enabling us to explore the impact of training with significantly larger volumes of data and its potential benefits for overall performance. English data is also included.
\end{itemize}

\begin{table}[]
\centering
\begin{tabular}{lccc}
\hline
\textbf{Configuration}  & \textbf{EU toks}  & \textbf{Avg.} \\ \hline
ZelaiHandi & 6.2B          & 61.22        \\
ZelaiItxi & 10.0B          & 61.71        \\ 
ZelaiHandi + Latxa & 13.8B        & 61.84        \\ \hline
\end{tabular}
\caption{\label{tab:cpt_results_datasets}Average performance results of continual pre-training using different datasets for Basque model adaptation. We report the number of Basque tokens seen during the continual pre-training phase.}
\end{table}

In Table \ref{tab:cpt_results_datasets}, we report the outcomes of our experiments on continual pre-training with different dataset configuration for Basque language model adaptation. The results reveal that while the largest dataset configuration—combining ZelaiHandi and Latxa—achieved the highest score (61.84), the improvement was only marginal compared to the smaller, high-quality datasets. ZelaiHandi and ZelaiItxi, despite their smaller size, provided competitive results (61.22 and 61.71, respectively). This suggests that the addition of significantly larger, noisier data from Latxa can slightly enhance performance, but high-quality, smaller datasets like ZelaiHandi offer a better balance between performance and computational efficiency. Additionally, while ZelaiItxi offers a larger dataset by extending ZelaiHandi with proprietary sources, its closed nature restricts broader usability and replicability, making ZelaiHandi more suitable for this work despite the smaller size.
Prioritizing high-quality and open license datasets proves to be a more practical and efficient approach for continual pre-training in resource-constrained environments.

\begin{table*}[t]
\centering
\begin{tabular}{l|ccc|ccc}
\hline
\textbf{Benchmark} & \makecell{\textbf{Latxa v1.2} \\ \textbf{7B}} & \makecell{\textbf{Llama 3.1} \\ \textbf{8B}} & \makecell{\textbf{Llama-eus} \\ \textbf{8B}} & \makecell{\textbf{Latxa v1.2} \\ \textbf{13B}} & \makecell{\textbf{Latxa v1.2} \\ \textbf{70B}} & \makecell{\textbf{Llama 3.1} \\ \textbf{70B}} \\ \hline
ARC\_eu         & 61.20 & \textbf{69.20} & 67.60 & 66.80 & 70.00 & \underline{\textbf{78.40}} \\ 
Winogrande\_eu  & 75.60 & \textbf{82.00} & 78.40 & 80.80 & 84.80 & \underline{\textbf{85.60}} \\ 
MMLU\_eu        & 38.15 & \textbf{66.67} & 62.59 & 47.41 & 51.48 & \underline{\textbf{72.22}} \\ 
HellaSwag\_eu   & 76.40 & \textbf{86.40} & \textbf{86.40} & 83.20 & 86.00 & \underline{\textbf{92.00}} \\ \hline
Belebele        & 41.56 & \textbf{87.44} & 84.67 & 63.44 & 81.78 & \underline{\textbf{94.44}} \\ 
X-StoryCloze    & 73.66 & 78.23 & \textbf{78.49} & 76.51 & 78.76 & \underline{\textbf{81.01}} \\ \hline
\rowcolor[HTML]{E6E6E6}\textbf{Average EN} & 61.10 & \textbf{78.32} & 76.36  & 69.69 & 75.47 & \underline{\textbf{83.95}} \\ \hline
\textbf{Average EU} & 53.14 & 52.06 & \textbf{63.08}  & 57.80 & 65.80 & \underline{\textbf{69.69}} \\
\textbf{Diff. (EN vs. EU)} & -9.96 & -26.26 & -13.28 & -11.89 & -9.67 & -14.26 \\ \hline
\end{tabular}
\caption{\label{tab:cpt_results_english}English evaluation results of models with fewer than 10B parameters and more than 10B parameters. \textbf{Bold} highlights the best results among models classified according to parameter counts, while the \underline{underlined} value denotes the overall best result across all configurations.}
\end{table*}

\section{Performance on English}
\label{app:english_performance}

We also assessed the English performance of our Llama-eus-8B model following the continual pre-training phase, as maintaining its initial competencies is crucial. The analysis revealed that the model experiences only a modest decrease of 1.96 points in average English scores compared to the base model, Llama-3.1-8B. This outcome indicates that while Llama-eus-8B has been effectively adapted for Basque, it continues to demonstrate a commendable level of competency in English, preserving its foundational knowledge. However, a significant performance gap, 13.28 points, remains between Basque and English across all evaluated models, highlighting the need for further enhancements in this area. Bridging this gap will be vital in future iterations, especially by leveraging transfer learning strategies to improve knowledge transfer from English to Basque. Table \ref{tab:cpt_results_english} shows additional details and insights into the results.

The results presented in Table \ref{tab:cpt_results_english} highlight the performance of the Llama-eus-8B model in comparison to both the Latxa and Llama models across various benchmarks. Notably, Llama-eus-8B demonstrates only a minor decrease in average English scores when compared to the base model Llama-3.1-8B, achieving a commendable average of 76.36 in English tasks. This suggests that while the model is effectively adapted for Basque, it retains a significant level of competency in English, thus preserving its foundational knowledge.

However, it is important to note the existing gap between Basque and English performance across all models. For instance, the average scores for English benchmarks are consistently higher than those for Basque, indicating that while Llama-eus-8B excels in Basque tasks, it has not fully bridged the performance gap between the two languages. Addressing this discrepancy will be crucial in future iterations of the model, particularly by leveraging transfer learning strategies that can facilitate better knowledge transfer from English to Basque. By focusing on reducing this gap, we can further enhance the model's capabilities and performance in Basque language tasks while maintaining its strong English competencies.

\section{Instruction Tuning Datasets}
\label{app:instruction_datasets}

To manually assess the instruction-following capabilities of the instructed models, we selected a random sample of 100 instructions from the No\_Robots test set, which contains 500 instructions. These 100 instructions were then manually translated into Basque. Table \ref{tab:no_robots_test} displays the number of examples from each task included in the manual test set, while Table \ref{tab:no_robots_examples} provides one example per task from this set.

\begin{table}[h]
    \centering
    \begin{tabular}{ll}
        \hline
        \textbf{Category}  &  \textbf{\# Examples}  \\
        \hline
        Generation         &  25                    \\
        Brainstorming      &  15                    \\
        Chat               &  15                    \\
        Open QA            &  13                    \\
        Classification     &  12                    \\
        Closed QA          &  5                     \\
        Extraction         &  5                     \\
        Rewriting          &  5                     \\
        Summarization      &  5                     \\
        \textbf{Total}     &  \textbf{100}          \\
        \hline
    \end{tabular}
    \caption{Number of examples per instruction category used in the manual evaluation of the instructed models.}
    \label{tab:no_robots_test}
\end{table}

\onecolumn

\begin{longtable}{|p{2.3cm}|p{6.2cm}|p{6.2cm}|}
    \hline
    \textbf{Category} &  \textbf{Original example}   &  \textbf{Translated example}  \\
    \hhline{|=|=|=|}
    Generation &  I need some product names.  These are nail polishes in various colors of pink and red. I have 15 that need names that sound kind of sexy and alluring. Alliteration is fine but they don’t always have to be alliterative.  Please keep the product name to two and three words each. Please number each.   &  Produktu izen batzuk behar ditut. Arrosa eta gorri kolore ezberdinetako azazkal-esmalteentzat dira. 15 izen sexy eta seduzitzaile dira behar ditudanak. Aliterazioa gustatzen zait, baina ez dute beti aliteratiboak izan behar. Bi edo hiru hitz izan behar ditu produktu bakoitzaren izenak. Mesedez, zenbakitu izen bakoitza.  \\
    \hline
    Brainstorming &  I'm about to plan a garden, it's going to be a flower garden, and I want to have a theme to it, but I can't think of any ideas that would fit a theme. I hope to plant several types of flowers, but I want them all the fit with the theme I choose, which is why I need to have a theme before I get them. Please help with this.   &  Lorategi bat antolatzekotan nago, lorez betetako lorategi bat izango da, eta gai baten ingurukoa izatea nahi dut, baina ez zait ezer bururatzen. Hainbat lore mota landatzea nahi dut, baina aukeratzen dudan gaiarekin bat etortzea nahi dut; horregatik, gaia pentsatuta izan behar dut loreak erosi aurretik. Mesedez, lagundu.  \\
    \hline
    Chat &  Gavin is a jealous chatbot that is often envious of users.\newline I want to travel to Bora Bora for vacation. Does it rain a lot in May?   &  Gavin txatbot jeloskorra da, eta askotan erabiltzaileen inbidia izaten du.\newline Bora Borara bidaiatu nahi dut oporretarako. Euri asko egiten du maiatzean?  \\
    \hline
    Open QA &  What is the International Atomic Time (TAI), and how does it differ from Coordinated Universal Time?   &  Zer da Nazioarteko Denbora Atomikoa (TAI) eta ze ezberdintasun du Denbora Unibertsal Koordinatuarekin?  \\
    \hline
    Classification &  What genres are these songs? Only list the genres, not the name of the song. If there are multiple genres, list those too.\newline \newline "Bohemian Rhapsody"\newline "Uptown Funk"\newline "Despacito"\newline "Someone Like You"\newline "Shape of You"\newline "Hotel California"   &  Zein generotakoak dira kantu hauek? Generoak bakarrik zerrendatu, ez abestien izenak. Genero bat baino gehiago baldin badago, aipa itzazu denak.\newline \newline "Bohemian Rhapsody"\newline "Uptown Funk"\newline "Despacito"\newline "Someone Like You"\newline "Shape of You"\newline "Hotel California"  \\
    \hline
    Closed QA &  How much has the City of Los Angeles spent fighting dust pollution from Owens Lake?\newline \newline Once a lake bed is exposed, winds kick up ferocious dust storms. Those windblown sediments contribute to air pollution and can contribute to asthma, lung cancer and cardiopulmonary disease, among other health issues. Owens Lake has been among the largest sources of dust pollution in the nation. The City of Los Angeles has spent more than \$2.5 billion mitigating the dust through projects at the lake bed such as shallow flooding, seeding and planting vegetation, spreading gravel or tilling the ground.   &  Zenbat gastatu du Los Angeles hiriak Owens lakuko hautsaren bidezko kutsadurari aurre egiten? \newline \newline Lakuaren hondoa agerian geratzen denean, haizeek hauts-ekaitz bortitzak sortzen dituzte. Haizeak eramandako sedimentu horiek airea kutsatzen dute, eta asma, biriketako minbizia eta bihotz-biriketako gaixotasunak eragin ditzakete, besteak beste. Owens lakua herrialdeko hauts-kutsaduraren iturri handienetako bat izan da. Los Angeles hiriak 2.500 milioi dolar baino gehiago gastatu ditu hautsa airetik kentzen, aintziraren hondoan proiektuak eginez, hala nola sakonera txikiko putzuak egiten, landaredia erein eta landatzen, legarra zabaltzen edo lurra lantzen.  \\
    \hline
    Extraction &  Please give me a numbered list extraction of all the sentences that don't have a citation.\newline \newline The total number of Greeks living outside Greece and Cyprus today is a contentious issue. Where census figures are available, they show around 3 million Greeks outside Greece and Cyprus. Estimates provided by the SAE - World Council of Hellenes Abroad put the figure at around 7 million worldwide.[187] According to George Prevelakis of Sorbonne University, the number is closer to just below 5 million.[186] Integration, intermarriage, and loss of the Greek language, influence the self-identification of the Greek diaspora (Omogenia); important centres include New York, Melbourne, London and Toronto.[186] In 2010, the Hellenic Parliament introduced a law that allowed members of the diaspora to vote in the Greek elections;[188] this law was later repealed in early 2014.[189]   &  Mesedez, eman iezadazu zenbakitutako zerrenda bat, aipurik ez duten esaldi guztiak aterata.\newline \newline Gaur egun Greziatik eta Zipretik kanpo bizi diren greziarren kopuru osoa gai polemikoa da. Zentsu-zifrek Grezia eta Zipretik kanpo 3 milioi greziar inguru daudela diote. SAEk (Atzerriko Heleniarren Munduko Kontseiluak) egindako kalkuluen arabera, 7 milioi inguru dira mundu osoan. [187] Sorbonneko Unibertsitateko George Prevelakisen arabera, zenbakia 5 milioitik gertuago dago. [186] Integrazioak, ezkontzak eta hizkuntza grekoaren galerak diaspora grekoaren (Omogenia) autoidentifikazioan eragiten dute; zentro garrantzitsu batzuk New York, Melbourne, Londres eta Toronto dira. 2010ean, Parlamentu heleniarrak diasporako kideei Greziako hauteskundeetan botoa ematea ahalbidetu zien lege bat sartu zuen;[188] lege hau 2014ko hasieran indargabetu zen. [189]  \\
    \hline
    Rewriting &  Rewrite this tweet to be kind and polite in its dissent.\newline \newline You’re literally a useful idiot to cover for fascists that push through policies that kill people   &  Berridatzi txio hau atsegina izan dadin, disidentea izanda ere.\newline \newline Pertsonak hiltzen dituzten politikak bultzatzen dituzten faxistak estaltzeko ergel erabilgarria zara, benetan.  \\
    \hline
    Summarization &  Tell me what this says in five words:\newline
    \newline It wasn’t long ago that record collecting seemed to be a niche hobby, indulged in by music lovers across the world, but hardly anyone else. Now, however, things have changed. Vinyl has seen a huge comeback – which may be surprising in the streaming age. Thanks to support from high-profile acts and “underground” artists alike, and major events like Record Store Day and Love Record Stores, the vinyl’s popularity has severely increased. So, if you’re a burgeoning vinyl lover wondering how to start a record collection, these six essential tips will get you up and running.   &  Esadazu zer kontatzen duen testu honek bost hitzetan:\newline \newline Ez da denbora asko pasa disko bilketak nitxoko denbora-pasa ematen zuenetik, mundu osoko musikazaleek egiten zutena, baina ia beste inork ez. Hala ere, gauzak aldatu egin dira. Biniloak bueltan dira, eta hori harrigarria izan daiteke streaming-aren aroan. Profil altuko ekitaldien eta “underground” artisten laguntzari esker, eta Record Store Day eta Love Record Stores bezalako ekitaldi garrantzitsuei esker, biniloaren ospeak gora egin du nabarmen. Beraz, binilo zale amorratua bazara eta disko bilduma bat nola hasi ez badakizu, oinarrizko sei aholku hauek zalantzak argituta eta hasteko prest utziko zaituzte.  \\
    \hline

    \caption{Examples of the instructions for the manual evaluation, one per category.}
    \label{tab:no_robots_examples}
\end{longtable}

\twocolumn

\begin{table*}[h!]
\centering
\begin{tabular}{|c|p{4cm}|p{4cm}|p{4cm}|}
\hline
\textbf{Lang.} & \textbf{Prompt} & \textbf{Chosen} & \textbf{Rejected} \\ \hline
EN & Classify the emotion conveyed in the following sentence as happy, sad, or neutral. ``I had a lovely day.'' & Sure! I would classify the emotion conveyed in the sentence ``I had a lovely day'' as happy. The word ``lovely'' implies a positive and enjoyable experience, which is typically associated with happiness. & Based on the context of the sentence, I would classify the emotion conveyed as neutral. The sentence contains no explicit mention of an emotion, and the context does not indicate otherwise. \\ \hline
EU (MT) & Sailkatu hurrengo otoitzean transmititutako emozioa zoriontsu, triste edo neutro gisa. ``Egun zoragarria izan nuen.'' & Bai horixe! ``Egun zoragarria izan nuen'' esaldian transmititutako emozioa zoriontsutzat sailkatuko nuke. ``Xarmangarri'' hitzak esperientzia positibo eta atsegin bat inplikatzen du, normalean zoriontasunarekin lotzen dena. & Otoitzaren testuinguruan oinarrituz, transmititutako emozioa neutrotzat sailkatuko nuke. Otoitzak ez du emozio baten aipamen espliziturik, eta testuinguruak ez du besterik adierazten. \\ \hline
\end{tabular}
\caption{Example of a triplet from the preference dataset UltraFeedback in English and Basque.}
\label{tab:preference_example}
\end{table*}

\section{Preference-tuning datasets}
\label{app:preferences_datasets}

UltraFeedback is a large-scale preference dataset comprising 256,000 samples generated by various LLMs, annotated by GPT-4 to evaluate four key aspects: instruction-following, truthfulness, honesty, and helpfulness, using prompts from diverse sources. Following the methodology described for the instruction datasets (see Section \ref{sec:ift_datasets}), we translated this dataset into Basque. The translated dataset, called \textbf{UltraFeedback\_eu}, consists of 61,135 triplets, each containing a prompt, a preferred response, and a rejected response.

Table \ref{tab:preference_example} shows an example of a triplet in the preference-tuning dataset. Both the original example and the machine translated equivalent in Basque are shown.

\begin{table*}[t]
    \centering
    \begin{tabular}{lccc} \hline
        \textbf{Model} & \textbf{Size} & \makecell{\textbf{Time} \\ \textbf{(GPU Hours)}} & \makecell{\textbf{Carbon emitted} \\ \textbf{(kg CO2 eq)}} \\ \hline
        Llama-eus-8B & 8B & 561.40 & 97.01 \\
        Llama-eus-8B-instruct & 8B & 199.76 & 34.52 \\
        Llama-eus-8B-instruct-DPO & 8B & 74.73 & 12.91 \\ \hline
        Total & - & 835.89 & 144.44 \\ \hline
    \end{tabular}
    \caption{Carbon footprint of training different models}
    \label{tab:carbon-emissions}
\end{table*}

\section{Carbon emmissions}
\label{app:carbon-emissions}

Adapting large language models (LLMs) to new languages involves compute-intensive experiments that contribute significantly to carbon emissions.To maximize efficiency, we focused on resource-efficient models, utilizing 8 billion parameter architectures along with techniques such as LoRA \cite{LoRA} and Flash Attention \cite{dao2022flashattention} to optimize computational performance.

All models were trained on eight NVIDIA A100 80GB SXM4 GPUs at the Donostia International Physics Center (DIPC) in Spain. We provide details on model size, compute hours, and carbon emissions for our experiments in Table \ref{tab:carbon-emissions}. Carbon emissions were estimated using the Machine Learning Impact calculator\footnote{https://mlco2.github.io/impact\#compute} developed by \newcite{lacoste2019quantifying}.

\end{document}